\PassOptionsToPackage{dvipsnames}{xcolor}
\documentclass[12pt, a4paper]{article}

\usepackage[utf8]{inputenc}

\usepackage[export]{adjustbox}
\usepackage{algorithm}
\usepackage[noend]{algpseudocode}
\usepackage{amsmath}
\usepackage{amssymb}
\usepackage{animate}
\usepackage{appendix}
\usepackage{array}
\usepackage{authblk}
\usepackage{bm}
\usepackage{booktabs}
\usepackage{caption}
\usepackage{color}
\usepackage{colortbl}
\usepackage{changepage}
\usepackage{chemformula}
\usepackage{comment}
\usepackage{csvsimple}
\usepackage{etoolbox}
\usepackage{enumitem}
\usepackage{float}
\usepackage{framed}
\usepackage{fontawesome5}
\usepackage[margin=1in]{geometry}
\usepackage{graphicx}
\definecolor{links}{HTML}{0074CC} % used to be 3366CC
\usepackage[colorlinks=true, allcolors=links]{hyperref}
\usepackage{lineno}
\usepackage{longtable}
\usepackage{lscape}
\usepackage{listings}
\usepackage{makecell}
\usepackage{marginnote}
\usepackage{mathrsfs}
\usepackage{mathtools}
\usepackage{mdframed}
\usepackage{multicol}
\usepackage{multirow}
\usepackage[numbers]{natbib}
\usepackage{pdflscape}
\usepackage{rotating}
\usepackage{tabulary}
\usepackage{tikz}
\usepackage[textsize=scriptsize]{todonotes}
\usepackage{setspace}
\usepackage{soul}
\usepackage{subcaption}
\usetikzlibrary{shapes, arrows, positioning, fit}
\usepackage{tabularray}
\usepackage{wrapfig}
\usepackage{xargs}
\usepackage[normalem]{ulem}
\usepackage{epstopdf}
\AppendGraphicsExtensions{.tif}

\PassOptionsToPackage{hyphens}{url}\usepackage{hyperref}

\definecolor{sandstorm}{rgb}{0.93, 0.84, 0.25}

\providecommand{\keywords}[1]{\textbf{{Keywords:}} #1}

\newcommand{\instructions}[1]{}

\newcolumntype{C}[1]{>{\centering\arraybackslash}p{#1}}

\providecommand{\keywords}[1]
{
  \noindent \small	
  \textbf{Keywords: } #1
}

\makeatletter
\newcommand*{\centerfloat}{%
  \parindent \z@
  \leftskip \z@ \@plus 1fil \@minus \textwidth
  \rightskip\leftskip
  \parfillskip \z@skip}
\makeatother

\AtBeginEnvironment{tabular}{\footnotesize} % Makes all tables at footnote size
\definecolor{miamired}{HTML}{C3142D}

\definecolor{cb1}{RGB}{166, 206, 227}  % Light blue
\definecolor{cb2}{RGB}{31, 120, 180}   % Dark blue
\definecolor{cb3}{RGB}{178, 223, 138}  % Light green
\definecolor{cb4}{RGB}{51, 160, 44}    % Dark green
\definecolor{cb5}{RGB}{251, 154, 153}  % Light red
\definecolor{cb6}{RGB}{227, 26, 28}    % Dark red
\definecolor{cb7}{RGB}{253, 191, 111}  % Light orange
\definecolor{cb8}{RGB}{255, 127, 0}    % Dark orange
\definecolor{cb9}{RGB}{202, 178, 214}  % Light purple
\definecolor{cb10}{RGB}{106, 61, 154}  % Dark purple
\definecolor{cb11}{RGB}{255, 255, 153} % Light yellow
\definecolor{cb12}{RGB}{177, 89, 40}   % Brown 
\definecolor{cb13}{RGB}{220, 220, 220} % Light gray
\definecolor{cb14}{RGB}{80, 80, 80}    % Dark gray

\usetikzlibrary{positioning,shapes,arrows,shadows,fit,backgrounds}

\title{\Large \textbf{Reliable Decision Support with LLMs: A Framework for Evaluating Consistency in Binary Text Classification Applications}}

\author[1]{Fadel M. Megahed}
\author[2]{Ying-Ju Chen}
\author[1]{L. Allison Jones-Farmer}
\author[1,*]{Gabe Lee}
\author[1]{Jiawei Brooke Wang}
\author[3,*]{Inez M. Zwetsloot}

\affil[1]{Farmer School of Business, Miami University, Oxford, OH 45056, USA}
\affil[2]{College of Arts and Sciences, University of Dayton, Dayton, OH 45469, USA}
\affil[3]{Amsterdam Business School, University of Amsterdam, Amsterdam, The Netherlands}
\affil[*]{Corresponding authors. Emails: \href{mailto:gabelee@miamioh.edu}{gabelee@miamioh.edu}, \href{mailto:i.m.zwetsloot@uva.nl}{i.m.zwetsloot@uva.nl}}

\date{\small \today}

\begin{document}

\maketitle

% {
% \textcolor{red}{Temporary author order: I just made it alphabetically after me and Tessa.} \newline \newline \textcolor{blue}{Not sure who wants to handle the submission? I have put Inez as a placeholder based on our last submission for JQT. But, we can figure out this part later if the group wants to have e.g., Gabe to be the corresponding author as he is a senior editor for DSS.} \newline \newline  \hl{(DSS has a strict 34-page limit so when you edit, add something, please remove something from the text to ensure that we are within the 34-page limit)}
% }

\begin{abstract}
\noindent This study introduces a framework for evaluating consistency in large language model (LLM) binary text classification, addressing the lack of established reliability assessment methods. Adapting psychometric principles, we determine sample size requirements, develop metrics for invalid responses, and evaluate intra- and inter-rater reliability. Our case study examines financial news sentiment classification across 14 LLMs (including \texttt{claude-3-7-sonnet}, \texttt{gpt-4o}, \texttt{deepseek-r1}, \texttt{gemma3}, \texttt{llama3.2}, \texttt{phi4}, and \texttt{command-r-plus}), with five replicates per model on 1,350 articles. Models demonstrated high intra-rater consistency, achieving perfect agreement on 90–98\% of examples, with minimal differences between expensive and economical models from the same families. When validated against \textit{StockNewsAPI} labels, models achieved strong performance (accuracy 0.76–0.88), with smaller models like \texttt{gemma3:1B}, \texttt{llama3.2:3B}, and \texttt{claude-3-5-haiku} outperforming larger counterparts. All models performed at chance when predicting actual market movements, indicating task constraints rather than model limitations. Our framework provides systematic guidance for LLM selection, sample size planning, and reliability assessment, enabling organizations to optimize resources for classification tasks.

% \vspace{2\baselineskip}

% \noindent \hl{Please read the potential contributions to the paper on pages 2-3.}

\textcolor{white}{  } % Just to add a line between abstract and keywords for readability
\end{abstract}

% added the three most relevant SDGs per their instructions (see choosing keywords section):
% https://authorservices.taylorandfrancis.com/publishing-your-research/writing-your-paper/using-keywords-to-write-title-and-abstract/


\newpage 
\maketitle

\newpage

\section{Introduction}
\label{sec:intro}

\begin{quote}
    \textit{LLM-based annotation has become something of an academic Wild West: the lack of established practices and standards has led to concerns about the quality and validity of research. Researchers have warned that the ostensible simplicity of LLMs can be misleading, as they are prone to bias, misunderstandings, and unreliable results} \citep[p.1]{tornberg2024best}.
\end{quote}
\begin{quote}

\textit{LLMs outperform typical human annotators. The evidence is consistent across different types of texts and time periods. It strongly suggests that ChatGPT may already be a superior approach compared to crowd annotations on platforms such as MTurk. At the very least, the findings demonstrate the importance of studying the text-annotation properties and capabilities of LLMs more in depth} \citep[p.2]{gilardi2023chatgpt}.

\end{quote}

\noindent Together, these contrasting perspectives highlight the need to critically examine large language models (LLMs) for text annotation and classification. Although human annotation remains widespread, it poses considerable challenges. It is time-consuming and costly—up to \$5 per annotation and \$50 per hour for annotators \citep{labelers2025explore}—and often suffers from inconsistencies stemming from the intricacies of language and the subjectivity of annotators \citep{wu2024modelling}. Differences in expertise, perspectives, and biases further affect the quality and reliability of human-labeled data \citep{davani2022dealing}.

Recent studies suggest that LLMs may address many of these limitations. Evidence shows that LLMs often outperform human annotators in tasks such as relevance assessment, position detection, and topic classification \citep{gilardi2023chatgpt, aguda2024large, chiang2023can, huang2023chatgpt, thapa2023humans, tornberg2023chatgpt}. Tools like \texttt{GPT-3} and \texttt{ChatGPT} have been shown to reduce annotation costs by 50\% to 96\%, driven by lower labor needs, faster task completion, and greater scalability \citep{gilardi2023chatgpt, wang2021want}. Meanwhile, the rapid adoption of Artificial Intelligence (AI) continues to drive demand for high-quality annotated data. Kniazieva \cite{kniazieva2024latest} projects that the U.S. data annotation market will grow by 25\% annually, reaching \$8.2 billion by 2028. Given this expansion and the persistent challenges of manual annotation, LLMs may substantially disrupt traditional practices.

Despite LLMs' promise for text annotation, preliminary studies highlight concerns about consistency, reliability, and validity. Evaluating summaries of 100 news articles by 16 LLMs, Stureborg et al. \cite[p.1]{stureborg2024large} found LLMs to be ``biased evaluators,'' exhibiting familiarity bias, uneven rating distributions, and anchoring effects. They also noted low inter-sample agreement and sensitivity to minor prompt variations. Aguda et al. \cite{aguda2024large} similarly identified a trade-off between LLMs' efficiency and the precision achieved by experts. Reiss \cite{reiss2023testing} urged caution in using LLMs for classification, emphasizing the need for validity studies. Collectively, the literature highlights the importance of structured methods for effectively integrating LLMs into annotation workflows \citep{tornberg2024best, pangakis2023automated}. 

In response to these concerns about consistency, reliability, and validity, this study proposes a structured framework for LLM-based text annotation and classification, designed to enhance scientific rigor and reproducibility. To inform the framework development, we examined four research questions:
\begin{enumerate}[label=(\arabic*), nosep]
    \item \textit{How can a statistical experiment be designed to measure the intra- and inter-rater reliability of LLM-based text annotation?} 
    % Unlike public-facing chat interfaces, programmatic access to LLMs enables precise control over several factors that may influence the model’s outputs, such as the model version (snapshot date), system prompt, temperature, maximum token limit, etc. Therefore, a controlled statistical experiment can systematically vary some of these parameters while holding the remaining constant. This would allow for identifying sources of variability and quantifying the consistency of the model’s annotations across repeated runs (intra-rater reliability) and across different models or configurations (inter-rater reliability). 

    \item \textit{What methods are appropriate for evaluating intra- and inter-rater reliability when multiple LLMs are involved in text classification?} 
    % Low intra-rater reliability indicates that the LLM's outputs are inconsistent, making it unreliable for the task. On the other hand, in the absence of a ``gold standard'' or a ``ground truth'' high agreement among LLMs can suggest that the classifications are relatively robust. 

    \item \textit{How can researchers determine the minimum sample size and number of replicates needed to reliably estimate LLM consistency?} 
    % In this context, sample size refers to the number of unique text items annotated, while replicates refer to the number of times each item is annotated. Determining appropriate values for both is critical for ensuring statistically robust estimates of reliability metrics (e.g., intra- and inter-rater agreement) while balancing the computational cost of large-scale experimentation.

    \item \textit{How to evaluate validity when multiple LLMs are involved in text classification?}
    %\textit{Given an established external criterion, can LLMs from different providers exhibit high inter-rater reliability while demonstrating low validity relative to the external criterion?} 
    % This scenario would suggest that while the LLMs agree amongst themselves, they may be consistently wrong or biased similarly, highlighting a potential systematic flaw in their annotation process and/or the task at hand.
\end{enumerate}

Guided by these research questions, we propose a structured framework to support reliable LLM-based annotation. The framework provides guidance on sample size planning, prompt design, model selection criteria, and a methodology to assess reliability and validity \citep{alizadeh2024open}. By following these procedures, researchers can evaluate the credibility and applicability of LLM-generated annotations \citep{liu2024poliprompt, nie2024survey}.  A financial news sentiment case study demonstrates the utility of the proposed framework.  Additionally, we provide open-source code and implementation guidance for broader adoption \citep{liu2024poliprompt, nie2024survey}.

\section{Background and Related Work}
\label{sec:relwork}
\subsection{Large Language Models}

Current LLMs are based on the transformer architecture, which efficiently processes large text sequences using self-attention. LLMs can be categorized as (a) encoder-only models like \texttt{BERT} for information extraction \citep{devlin2019bert}; (b) decoder-only models like \texttt{GPT} for fluent text generation \citep{brown2020language}; or (c) encoder-decoder models like \texttt{BART} for summarization and translation \citep{lewis2020bart}. LLM outputs are inherently stochastic, with repeated prompts yielding different results.

As LLMs have scaled, they exhibit emergent abilities such as solving math problems and understanding ambiguous language \citep{wei2022characterizing}. These capabilities support their classification as foundation models \citep{bommasani2022opportunities}, enabling broad application with minimal customization across tasks like text retrieval \citep{shi2025know}, entity matching \citep{babaian2024entity}, sentiment analysis \citep{mousavi2024resilience}, and text classification \citep{sun2023text}—the focus of this study. LLMs are increasingly regarded as general-purpose technologies \citep{eloundou2024gpts}, disrupting labor markets \citep{wef2025future, handa2024economic, demirci2025ai} and firm performance \citep{dell2023navigating, eisfeldt2023generative, reusens2025llm}. However, fundamental uncertainties about their capabilities and failure modes persist \citep{salazar2025contribution, bommasani2022opportunities, eloundou2024gpts}.

\subsection{LLMs for Text Classification / LLMs-as-a-Judge}

Since the release of \texttt{ChatGPT}, LLMs have seen growing use in text classification \citep{brown2020language, ouyang2022training}. Two main factors explain this shift: (1) LLMs can operate in zero- or few-shot settings, reducing reliance on labeled datasets and feature engineering \citep{brown2020language, chae2023large}; and (2) NLP methods are foundational across fields such as finance, information systems, and accounting \citep{maibaum2024selecting, mousavi2024resilience, shao2025revisiting}. Comparative studies show that \texttt{GPT}-based models often outperform fine-tuned \texttt{BERT} models, with strong generalization and minimal tuning \citep{pangakis2023automated, maibaum2024selecting, peeters2023entity, liu2023pre}. However, computational cost, interpretability, and prompt sensitivity remain challenges \citep{tornberg2024best}.

Prompt engineering significantly affects classification outcomes \citep{sun2023text, chae2023large, peeters2023entity}. Chain-of-Thought (CoT) prompting, which guides models through reasoning steps, includes zero-shot CoT (“think step by step”) and manual few-shot CoT with examples \citep{wei2022chain, Weng2023-az}. Techniques like role-playing can enhance CoT \citep{kostina2025large}. Manual CoT has been used for fake news detection, job review classification, and grading student writing \citep{kostina2025large, eloundou2024gpts, eisfeldt2023generative, impey2025using}, though no method consistently outperforms others across tasks \citep{tornberg2024best}.

A major gap in current research is inconsistent evaluation protocols. Studies often omit replicates, proper sample sizing, or power analysis. For example, Eisfeldt et al. \cite{eisfeldt2023generative} used 100 samples and three replicates without justification. Findings also conflict: Kostina et al. \cite[p. 5]{kostina2025large} found deterministic behavior at temperature 0, contradicting Eisfeldt et al. \cite{eisfeldt2023generative} and Megahed et al. \cite{megahed2024introducing}. Verga et al. \cite{verga2024replacing} proposed a diverse panel of models to reduce bias and cost but do not address replication.

\subsection{Intra and Inter-Rater Reliability}

Intra-rater reliability captures a single rater’s consistency across repeated assessments, while inter-rater reliability measures agreement across multiple raters \citep{krippendorff2019content, shrout1979intraclass}. Both are essential for trustworthy measurement, as inconsistency can obscure true values. These concepts are widely used across disciplines such as psychology, healthcare, communication, and NLP \citep{krippendorff2019content}. Several statistical measures quantify rater reliability, varying by data type (e.g., categorical vs. continuous) and rater configuration (e.g., two vs. many raters) \citep{gwet2021handbook}. Key metrics include:

\textbf{Simple Agreement} is the proportion of times the raters assign the same label. Although intuitive, it does not account for chance agreement.

\textbf{Cohen’s Kappa ($\kappa$)} \citep{cohen1960coefficient} adjusts for chance agreement between two raters on categorical data: $$\kappa=\frac{P_o-P_e}{1-P_e},$$
where $P_o$ is observed agreement and $P_e$ is expected agreement by chance. $\kappa$ ranges from $-1$ (systematic disagreement) to $1$ (perfect agreement). Landis and Koch \citep[p.165]{landis1977measurement} suggest that $\kappa$ values between 0.61 and 0.80 represent “substantial” agreement. 

\textbf{Fleiss’ Kappa} \citep{fleiss1971measuring} extends Cohen’s kappa to multiple raters ($m \geq 3$), assuming the same set of raters rates all items. It provides a chance-corrected agreement measure for nominal data.

\textbf{Conger’s Kappa} \citep{conger1980integration} also generalizes Cohen’s kappa to multiple raters but allows for varying rater-item assignments.

\textbf{Brennan-Prediger Coefficient} \citep{brennan1981coefficient} uses an alternative approach to chance correction by assuming uniform category distribution. It is less sensitive to unbalanced marginal distributions than Cohen’s kappa but may still overestimate reliability when the class distributions are unbalanced.

\textbf{Krippendorff’s Alpha ($\alpha$)} \citep{krippendorff1970bivariate} is a general-purpose reliability measure that accommodates any number of raters and data types (nominal, ordinal, interval, ratio). It supports missing data and utilizes appropriate distance functions based on the data type. Hayes and Krippendorff \citep{hayes2007answering} advocate $\alpha$ as the standard for content analysis. T{\"o}rnberg \cite{tornberg2024best} applied $\alpha$ in ``prompt stability analysis'' for LLMs. 

\textbf{Gwet’s AC1} \citep{gwet2002inter} addresses paradoxes in kappa metrics under unbalanced class distributions. It assumes partial randomness in rater behavior and limits expected chance agreement to 50\% per category. While robust to prevalence effects, AC1 may overestimate reliability in balanced datasets \citep{gwet2008computing}.

\subsection{Sample Size Calculations}

Many foundational reliability measures originated in behavioral sciences, where sample size planning is essential due to the high cost of collecting responses or ratings \citep{krippendorff2019content, gwet2021handbook}. Researchers routinely estimate the minimum sample size needed for stable reliability estimates, particularly when assessing internal consistency or rater agreement. In contrast, LLM-based classification tasks often involve massive corpora. For example, Sutherland et al. \cite{sutherland2024occupational} analyzed over 530,000 CPA profiles, illustrating the scale of LLM-assisted annotation. Downsampling enables cost-effective, statistically grounded evaluations of model reliability and validity. Using a smaller, representative sample avoids wasting resources on unreliable models. Once a model demonstrates sufficient consistency and validity, it can be confidently applied to the full dataset. Our approach supports this workflow by providing statistical tools to estimate sample sizes needed for robust reliability assessment.

%Text classification tasks in both research and industry often involve massive datasets, where automated LLM-based annotation can offer substantial value. For example, Sutherland et al. \cite{sutherland2024occupational} analyzed profiles of over 530,000 certified public accountants, highlighting the scale at which LLM-assisted annotation could be deployed. Our approach proposes comparing models on a representative sample and provides statistical calculations to guide users in selecting appropriate sample sizes to ensure satisfactory power across different reliability metrics. 

\section{Our Proposed Framework}
\label{sec:method}
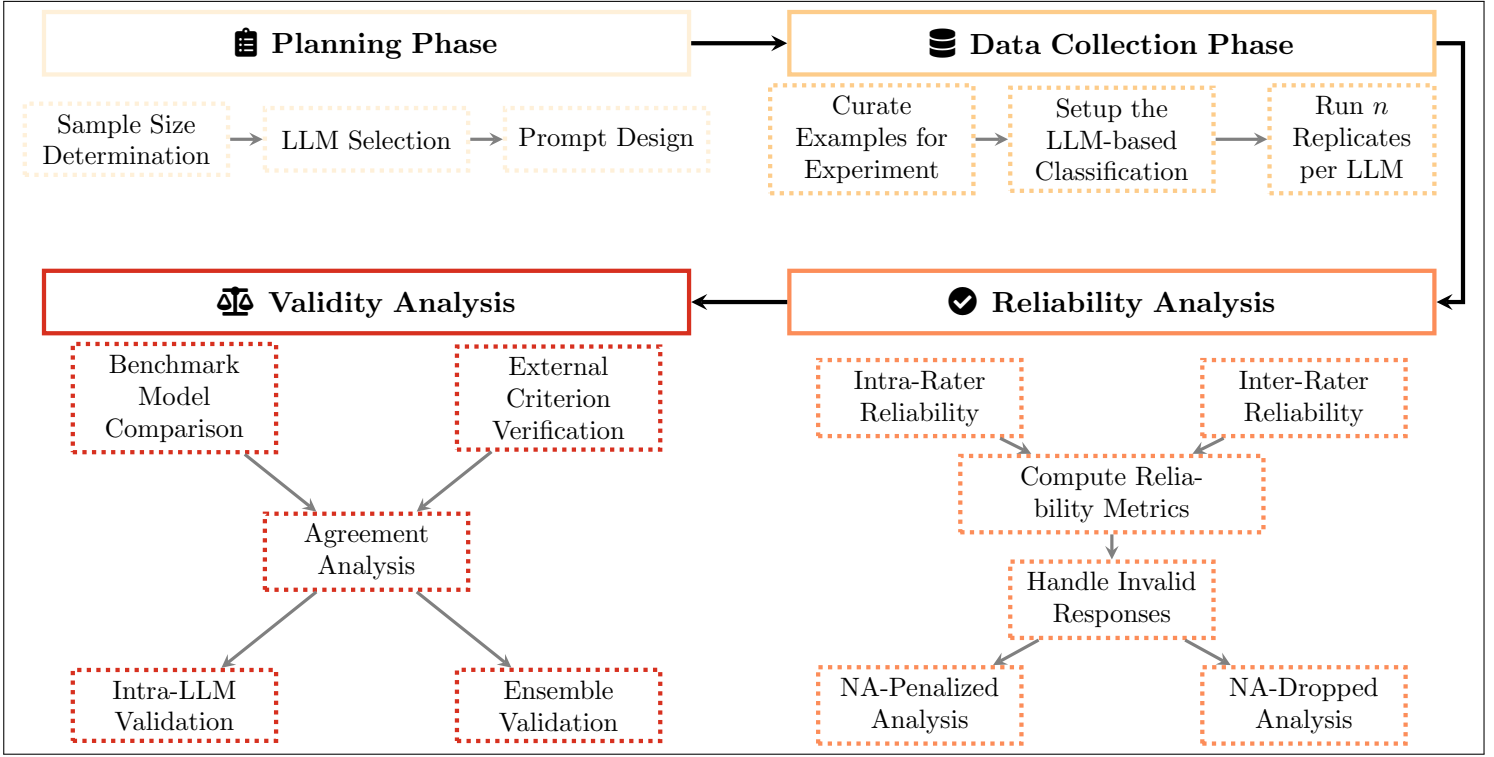
\begin{figure}[htb!]
    \centering
    \begin{adjustwidth}{-0.65in}{-0.65in}
    \fbox{
        \begin{tikzpicture}[node distance=0.5in and 1in, font=\footnotesize]

        % Define styles specific to this TikZ figure
        \tikzset{
            phase/.style = {rectangle, draw=black, thick, solid, ultra thick, text width=5.5cm, minimum height=2em, text centered, fill=white, text width=3.25in, font=\small},
            subcomponent/.style = {rectangle, draw=black, dotted, ultra thick, minimum height=2em, text centered, fill=white, text width = .95in},
            arrow/.style = {very thick,->,>=stealth, gray},
            arrow2/.style = {ultra thick,->,>=stealth},
            color1/.style={draw={rgb,255:red,254;green,240;blue,217}, fill=white},
            color2/.style={draw={rgb,255:red,253;green,204;blue,138}, fill=white},
            color3/.style={draw={rgb,255:red,252;green,141;blue,89}, fill=white},
            color4/.style={draw={rgb,255:red,215;green,48;blue,31}, fill=white}
        }
        
        % Planning Phase
        \node (planning) [phase, color1] {\faClipboardList \, \textbf{Planning Phase}};
        \node (llm_selection) [subcomponent, below of=planning, color1] {LLM Selection};
        \node (sample_size) [subcomponent, left of=llm_selection, xshift=-.75in, color1] {Sample Size Determination};
        \node (prompt_design) [subcomponent, right of=llm_selection, xshift=.75in, color1] {Prompt Design};
        
        \draw[arrow] (sample_size) -- (llm_selection);
        \draw[arrow] (llm_selection) -- (prompt_design);
        
        % Data Collection Phase
        \node (data) [phase, right=0.5in of planning, color2] {\faDatabase \, \textbf{Data Collection Phase}};
        \node (systematic_class) [subcomponent, below of=data, color2] {Setup the LLM-based Classification};
        \node (text_input) [subcomponent, left of=systematic_class, xshift=-0.75in, color2] {Curate Examples for Experiment};
        \node (multiple_replicates) [subcomponent, right of=systematic_class, xshift=0.75in, color2, text width=.75in] {Run $n$ Replicates per LLM};
        
        \draw[arrow] (text_input) -- (systematic_class);
        \draw[arrow] (systematic_class) -- (multiple_replicates);
        
        % Reliability Analysis
        \node (reliability) [phase, below=1in of data, color3] {\faCheckCircle \, \textbf{Reliability Analysis}};
        \node (intra_rater) [subcomponent, below of=reliability, xshift=-1in, color3] {Intra-Rater Reliability};
        \node (inter_rater) [subcomponent, below of=reliability, xshift=1in, color3] {Inter-Rater Reliability};
        \node (compute_metrics) [subcomponent, below of=reliability, yshift=-0.5in, text width = 1.47in, color3] {Compute Reliability Metrics};
        \node (invalid_responses) [subcomponent, below of=compute_metrics, yshift=-0.05in, color3] {Handle Invalid Responses};
        \node (na_penalized) [subcomponent, below of=invalid_responses, xshift=-1in, yshift=-0.05in, color3] {NA-Penalized Analysis};
        \node (na_dropped) [subcomponent, below of=invalid_responses, xshift=1in, yshift=-0.05in, color3] {NA-Dropped Analysis};
        
        \draw[arrow] (intra_rater) -- (compute_metrics);
        \draw[arrow] (inter_rater) -- (compute_metrics);
        \draw[arrow] (compute_metrics) -- (invalid_responses);
        \draw[arrow] (invalid_responses) -- (na_penalized);
        \draw[arrow] (invalid_responses) -- (na_dropped);
        
        % Validity Analysis
        \node (validity) [phase, below=1in of planning, color4] {\faBalanceScale \, \textbf{Validity Analysis}};
        \node (gold_standard) [subcomponent, below of=validity, xshift=-1in, color4] {Benchmark Model \\Comparison};
        \node (ground_truth) [subcomponent, below of=validity, xshift=1in, color4] {External Criterion Verification};
        \node (agreement_analysis) [subcomponent, below of=validity, yshift=-0.8in, color4] {Agreement Analysis};
        \node (individual_llm) [subcomponent, below of=agreement_analysis, xshift=-1in, yshift =-0.3in, color4] {Intra-LLM Validation};
        \node (ensemble_validation) [subcomponent, below of=agreement_analysis, xshift=1in, yshift =-0.3in, color4] {Ensemble Validation};
        
        \draw[arrow] (gold_standard) -- (agreement_analysis);
        \draw[arrow] (ground_truth) -- (agreement_analysis);
        \draw[arrow] (agreement_analysis) -- (individual_llm);
        \draw[arrow] (agreement_analysis) -- (ensemble_validation);
        
        % Phase connections
        \draw[arrow2] (planning) -- (data);
        \draw[arrow2] (data.east) -- ++(.35,0) |- (reliability.east);
        \draw[arrow2] (reliability) -- (validity);
        
        \end{tikzpicture}
        }
    \end{adjustwidth}
    \caption{Our framework for evaluating LLM consistency in binary text annotation tasks.}
    \label{fig:framework}
\end{figure}

Our framework for evaluating LLM-based text classification includes four phases (Figure \ref{fig:framework}): planning, data collection, reliability analysis, and validity analysis. The planning phase determines the minimum sample size, selects LLMs, and designs prompts. Data collection involves curating examples, implementing code infrastructure, and running the annotation experiment. Reliability analysis assesses annotation consistency across runs, while validity analysis measures accuracy against benchmark or external criterion labels. The framework is designed to be domain-agnostic and can be applied to any binary classification task.

\subsection{Planning Phase}

Evaluating the reliability of binary text annotation requires careful planning due to the complexity of the task. We recommend computing the following complementary reliability metrics: \texttt{Conger's Kappa}, \texttt{Fleiss' Kappa},  \texttt{Gwet's AC1}, \texttt{Brennan-Prediger Coefficient}, and \texttt{Krippendorff's Alpha}. 
% \texttt{Conger's Kappa} and \texttt{Fleiss' Kappa} extend Cohen's Kappa to multi-rater settings, offering robust adjustments for chance agreement, which refers to the level of agreement that could occur simply due to random factors. The \texttt{Simple Percent Agreement} provides an intuitive baseline measure of consistency, while \texttt{Gwet's AC1} addresses potential biases in Kappa statistics, such as underestimating agreement when category prevalence is unbalanced. The \texttt{Brennan-Prediger Coefficient} is especially effective for binary outcomes, offering a simple yet robust adjustment for chance agreement in cases of extreme prevalence. \texttt{Krippendorff's Alpha} has been recommended for LLM reliability analysis by \citet{tornberg2024best}. 
Using multiple metrics ensures a more comprehensive evaluation, as no single measure fully captures the nuances of reliability in binary annotation. These metrics can be efficiently computed using open-source tools such as the \texttt{irrCAC} package \citep{gwet2019irrCAC}, enabling reproducible and systematic assessment for similar tasks.

\subsubsection{Sample Size Determination}
\label{frame:sample}

To determine the minimum required examples, we adapted Gwet's \cite{gwet2021handbook} psychometric approach. The examples to be annotated are treated as analogous to ``subjects,'' while the ``raters'' represent either multiple LLMs (inter-reliability) or multiple replicates of a single LLM's outputs (intra-reliability). 
% Our study focused on the intra-reliability case, using replicates from a single LLM to capture variability in annotation performance. This approach minimizes computational costs and prepares for a worst-case scenario where multiple LLMs may be required, corresponding to an inter-reliability setup. By ensuring sufficient replicates ($r=5$), the framework provides enough data to evaluate reliability across both scenarios while maintaining scalability. 

We selected a target margin of error ($E_0 = 0.10$) and a confidence level of ($1-\alpha = 0.90$), standard values in psychometric research. From the psychometric literature, sample size estimation techniques are only presented for three reliability metrics: \texttt{Percent Agreement}, \texttt{Gwet’s AC1}, and \texttt{Brennan-Prediger Coefficient} \citep{gwet2021handbook}. For each metric, the sample size can be calculated as:

\begin{equation}
n_0 = \frac{z_\alpha^2 C}{E_0^2},
\end{equation}

\noindent where $z_\alpha$ is the critical value corresponding to the confidence level ($\approx1.645$ for 90\%), $C$ is a parameter that depends on the reliability metric, number of raters/replicates ($r$), and the number of categories ($q$), and $E_0$ is the desired margin of error. This ensures the sample size is large enough to achieve the desired level of precision by controlling the standard error of the agreement coefficient. The $C$ values can be computed as $C = \frac{1}{A}$, where $A$ is the agreement value for a given metric. Values of $A$ can be obtained from the agreement tables provided in Gwet's handbook \cite{gwet2021handbook}, using the number of categories ($q$) and the number of replicates/raters ($r$). The tables show that $A$ increases with $q$ and $r$. For binary classification, $q=2$ since the output is binary. Conversely, $r$ should be set to the number of planned replicates per model in the case of intra-rater reliability. %This choice will produce the largest (most conservative) sample size. 
By inserting the values for $C$, $E_0$, and $z_{\alpha}$, we obtain:

\begin{equation}\label{equ:sample_size}
n_{\text{PA}} = \frac{z_\alpha^2 C_{\text{PA}}}{E_0^2}, \quad
n_{\text{AC1}} = \frac{z_\alpha^2 C_{\text{AC1}}}{E_0^2}, \quad
n_{\text{BP}} = \frac{z_\alpha^2 C_{\text{BP}}}{E_0^2},
\end{equation}

\noindent where the subscripts $PA$, $AC1$, and $BP$ denote \texttt{Percent Agreement}, \texttt{Gwet’s AC1}, and \texttt{Brennan-Prediger}. A conservative final sample size can then be computed as: 
\begin{equation}
n_{\text{final}} = \max \left(n_{\text{PA}}, \ n_{\text{AC1}}, \ n_{\text{BP}}\right).
\end{equation}

\noindent Inter-rater reliability is computed by combining all replicates per LLM into a single rating set and assessing agreement across models without a separate sample size calculation. This adaptation of psychometric principles to LLM annotation tasks provides a systematic approach to determining the minimum number of examples required for reliability assessment.
% By focusing on intra-reliability with five replicates, we ensure a robust evaluation of annotation consistency while maintaining flexibility to extend the framework to inter-reliability scenarios.

\subsubsection{Selecting Candidate LLMs}\label{subsubsec:selecting_llms}

Selecting candidate LLMs requires careful consideration of several factors. First, data sensitivity may necessitate locally deployed models rather than API-based services. Second, computational capacity and associated costs can significantly impact the feasibility of large-scale annotation. Third, expected model performance should be evaluated using established benchmarks and leaderboards \citep{chiang2024chatbot}. Together, these considerations help identify models that align with task needs and practical constraints.

\subsubsection{Prompt Design and Engineering}

Prompt design for text annotation includes two key components \citep{tornberg2024best}: a detailed codebook and effective prompt engineering. The codebook should define categories and outline how to handle edge cases \citep{glaser1999discovery, tornberg2024best}. Prompt engineering should use techniques like \texttt{manual CoT} to elicit step-by-step reasoning \citep{amatriain2024prompt, minaee2024large}. Both the codebook and prompts should be iteratively refined through testing to ensure the outputs meet expectations.

\subsection{Data Collection Phase}

\subsubsection{Curating Representative Examples for Annotation}

The first step is to curate a dataset with at least $n_{\text{final}}$ examples, prioritizing diverse and edge cases relevant to the task \citep{tornberg2023chatgpt}. If no benchmark exists, a stratified random sample should be drawn from the full dataset. The curated set should be class-balanced and include both the text to be annotated and, when available, benchmark or external criterion labels.

% It is important to note that the distinction between ``gold standard'' and ``ground truth'' labels is not always clear-cut. For example, in sentiment analysis of subjective text, such as sarcastic tweets, an actual ``ground truth'' label may not exist due to varying interpretations. In such cases, the ``gold standard'', derived from expert annotations or benchmark models, serves as a substitute for the true ground truth, despite being subject to inherent biases or classification errors.

\subsubsection{Setting Up the LLM-based Classification System}

The second step is to develop and validate the code infrastructure for LLM-based classification. The infrastructure should support annotation, manage API or computational constraints, and ensure proper data management throughout the experiment. We recommend testing with a small subset ($\le 10$ examples) to verify functionality before running the full experiment.

The code consists of three primary components. First, we implement the data management and replication logic. This component transforms the curated dataset into a format suitable for repeated LLM annotations, tracking example-level information (e.g., timestamp, model, replicate). Unlike reliability studies with human annotators, LLM interactions through chat completion interfaces are memoryless, eliminating the need to randomize the run order. 
% This characteristic can simplify the experimental design but requires consideration of API rate limits and computational resources. 
% Hence, three implementation approaches are possible: (a) \textit{Sequential Replicate Completion}, which processes all replicates for each example-LLM pair before moving to the following example; (b) \textit{Batch Replicate Processing}, which completes one replicate for all examples before initiating the next replicate, which can be preferable for investigating temporal effects in API responses; and (c)  \textit{Fully Randomized Processing}, which ignores the memory-less property of programmatic interactions with LLM and randomly processes example-replicate-LLM combinations. 

Second, we developed the LLM interaction layer, which manages communication with the models. This includes a chat completion function that handles locally hosted and API-based model interactions while implementing retry mechanisms for failed requests. The function logs completions, manages rate limits, standardizes output formats, and ensures robust parsing to extract classifications from responses that may deviate from requested formats. % We recommend implementing robust parsing functions that extract the relevant classification while logging any unexpected response variations. Researchers and practitioners can utilize our code (see supplementary materials) or leverage novel open-source packages to facilitate these interactions \citep[e.g., see][]{ruiz2024mall, pieper2024batchllm}.

Third, we use the \texttt{irrCAC} package \citep{gwet2019irrCAC} to compute reliability metrics and agreement scores. This component must be designed to process invalid responses outside the prescribed annotation categories (e.g., ``neutral'' when only ``positive'' or ``negative'' options are specified).%, and calculating both standard reliability metrics and their LLM-inspired modifications. 
% While custom implementations of these components may be necessary for specific use cases, researchers can also leverage emerging packages such as \texttt{mall} \citep{ruiz2024mall} or \texttt{batchLLM} \citep{pieper2024batchllm} that provide specialized functionality for LLM experiments. These packages offer efficient methods for batch processing and result aggregation, though they should be evaluated for compatibility with the specific requirements of the annotation task. Although we did not use these packages in our case study—since they were released after our experiment was set up—researchers can use these \faRProject \ packages or utilize the Python code we provide in the supplementary materials to suit their needs.

The LLM-based classification system should be validated through a comprehensive testing phase that verifies the correct handling of requests and responses, proper implementation of retry logic/error handling, accurate parsing and storage of LLM outputs, appropriate computation of reliability metrics, and management of computational resources. Only after successfully validating these components should one proceed to a full-scale reliability experiment.

\subsubsection{Running \textit{n} Replicates per LLM}

The final step of the data collection phase involves executing the complete annotation reliability experiment, processing $n_{\text{final}} \times r \times m$ LLM calls (annotation tasks), where $n_{\text{final}}$ is the minimum sample size determined in the planning phase, $r$ is the number of replicates, and $m$ is the number of LLMs under evaluation. This step produces a comprehensive dataset containing the raw LLM outputs, parsed classifications, and relevant metadata.

\subsection{Reliability Analysis Phase}

The reliability analysis phase evaluates annotation consistency using intra-rater and inter-rater reliability approaches.
% In the intra-rater case, each replicate of a single LLM's annotations is treated as a distinct rater, allowing us to assess the consistency of a single model across multiple runs. For inter-rater reliability, each LLM is considered a distinct rater, enabling consistency evaluation across different models. 
% The computation of reliability metrics follows established psychometric practices when all annotations are valid. 
We employ five complementary reliability metrics: \texttt{Conger's Kappa}, \texttt{Fleiss' Kappa}, \texttt{Gwet's AC1}, \texttt{Brennan-Prediger Coefficient}, and \texttt{Krippendorff's Alpha}. Each metric offers unique insights into annotation consistency.
% , with computation handled through the \texttt{irrCAC} package \citep{gwet2019irrCAC}.

LLM-based annotation introduces unique challenges when models provide invalid or unexpected responses. We propose coding invalid annotations as \texttt{NA} and using two approaches to computing \texttt{simple percent agreement}: (a) \texttt{NA}-dropped analysis: excludes invalid responses from the analysis, providing an optimistic view of reliability by considering only cases where the LLM successfully produced valid classifications, and (b) \texttt{NA}-penalized analysis: treats invalid responses as disagreements, directly penalizing the LLM for failing to provide valid classifications.
% . Invalid responses can indicate an LLM's fundamental inability to handle specific annotation tasks, issues with prompt design, or a combination of both factors. 
% We propose two complementary approaches to computing simple percent agreement to address the invalid response issue while maintaining interpretability. The two approaches are:\todo{Tessa, please fix this since we will keep one of these. Is that correct?} 
% \begin{enumerate}[label=(\arabic*), nosep]
%     \item \textit{NA-Dropped Analysis}: Following traditional psychometric practice, this approach excludes invalid responses from the analysis. The exclusion is done case-wise. For example, if a given LLM presents 3 ``positive'', 1 invalid response, and 1 ``negative'' response, the dropped percent agreement is 75\% (3 out of 4 valid answers). This provides an optimistic view of reliability by considering only cases where the LLM successfully produced valid classifications. The analysis helps identify the model's consistency when it can complete the task as intended.
%     \item \textit{NA-Penalized Analysis}: This more conservative approach treats invalid responses as disagreements, directly penalizing the LLM for failing to provide valid classifications. When calculating agreement, invalid responses automatically count against consistency scores. For the example above, the penalized percent agreement in this case is 60\% (3 out of 5 total answers). This approach helps identify systematic issues in LLM performance or prompt design that might be masked by simply dropping invalid responses.
% \end{enumerate}
Both approaches can lead to ties, which do not impact the agreement calculation on the case/example level but will impact the label choice (and aggregate computations). We break ties by randomly assigning one of the two labels. Handling invalid responses for the remaining metrics follows standard psychometric practice, with exclusions determined independently for each metric based on its computational requirements. Specifically, \texttt{Fleiss' Kappa} and \texttt{Gwet's AC1} will return \texttt{NA} when all raters give invalid responses for a subject. However, the remaining metrics will still compute values. 
% This difference in behavior between metrics provides additional insight into the nature and impact of invalid responses in the annotation process.

% Combining these approaches, using multiple reliability metrics and two methods for handling invalid responses in the simple percent agreement, provides a comprehensive view of annotation consistency. This allows researchers to identify how reliable the annotations are when successful and how frequently and systematically LLMs fail to provide valid responses.

\subsection{Validity Analysis Phase}

The validity analysis phase evaluates the predictive (criterion-related) validity of LLM-generated sentiment annotations by comparing them against two reference standards: (1) a benchmark generated by a high-performing reference model and (2) an external criterion. We assess predictive validity at both the individual LLM level and the ensemble level. For individual LLMs, we evaluate annotations across replicates using both \texttt{NA}-dropped and \texttt{NA}-penalized approaches to handle missing data. The ensemble analysis aggregates annotations via majority voting across all LLMs (using their first replicate) to produce a single label per example, which helps address inconsistencies in individual model performance. In cases of ties during ensemble voting, one of the tied labels is randomly assigned. While ensembling can enhance annotation robustness by leveraging the collective outputs of multiple models, it may be unnecessary when individual models already exhibit high agreement.

% This idea parallels the principles of machine learning ensembling, where methods like bagging or boosting improve performance primarily for weak classifiers by reducing variance or bias. 

% , consistent with our approach to handling ties in reliability analysis.

\section{Example: Financial News Sentiment Annotation}
\label{sec:exp}

Financial news directly impacts investor sentiment and market behavior. Retail investors, institutional traders, and financial analysts rely on news sentiment to forecast stock movements, perform risk assessments, and develop trading strategies. Sentiment analysis enhances automated trading systems, strengthens risk management models, and provides early signals of market trends. LLMs represent a potentially transformative technology for sentiment classification that could result in automated financial systems with these models embedded in their decision processes. 

Our case study investigates binary sentiment classification of financial news using LLMs. Each model receives an article’s title, text, and ticker symbol and must predict whether the news will have a “positive” or “negative” effect on the stock’s next-day return. We focus on consistency: whether an LLM produces the same classification when analyzing identical inputs to the same model and across different models.

\subsection{Planning Phase}

\paragraph*{Sample size determination} Our case study implements the psychometric approach outlined in Section \ref{frame:sample} with a practical adjustment. We applied a \v{S}id\'{a}k correction to control the family-wise error rate when conducting multiple comparisons. We adjusted the base significance level $\alpha$ to $\alpha^* = (1-0.9^{1/7}) \approx 0.0149$, where 0.90 represents our target confidence level, and 7 represents the maximum number of LLMs we would realistically explore in the intra-rater reliability scenario. We chose seven as a conservative upper bound on the number of LLMs likely to pass the intra-rater reliability step. This ensures that the adjusted significance level reflects the maximum set of pairwise comparisons that practitioners would reasonably conduct in this phase. This adjustment produces a more conservative critical value for $z_{\alpha^*/2}$ to compensate for the increased risk of type I error in multiple model comparisons.

\begin{figure}[htb!]
\centering
\includegraphics[width=0.99\textwidth,frame]{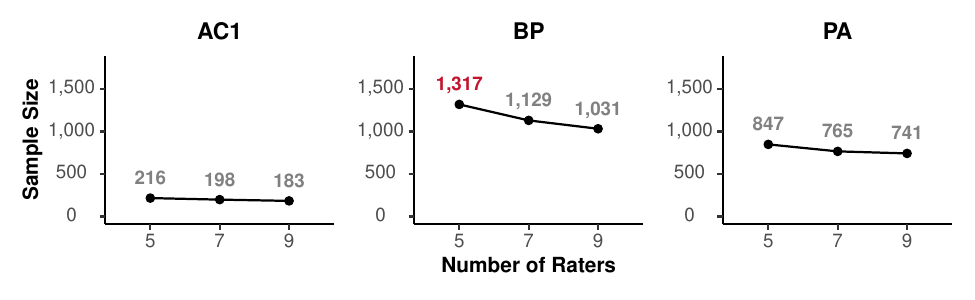}
\caption{Minimum sample sizes needed for the \texttt{Gwet AC1}, \texttt{Brennan-Prediger coefficient} (BP), and \texttt{simple percent agreement} (PA) for 5, 7, and 9 raters. The red value denotes the conservative (maximum sample size) across the three metrics.}
\label{fig}
\end{figure}

We extracted inverse $C$ values from Gwet’s agreement tables \cite{gwet2021handbook} based on binary classification ($q=2$) and rater counts ($r=5, 7, 9$). We used odd values of $r$ to reduce the likelihood of ties.  Figure~\ref{fig} displays the resulting sample sizes across reliability metrics. \texttt{Brennan-Prediger} required the largest sample ($n_{final} = 1{,}317$), \texttt{Percent Agreement} a moderate size (847), and \texttt{Gwet’s AC1} the smallest (216). All metrics show reduced sample size needs as the number of raters increases, reflecting the statistical efficiency of additional raters.

We adopted a conservative sample size strategy, selecting five replicates for the intra-rater evaluation to provide meaningful granularity in agreement measures (allowing for 60\%, 80\%, or 100\% agreement) while maintaining computational feasibility. This choice balances the minimal three replicates and the upper bound of ten typically used in psychometric studies. We applied the \v{S}id\'{a}k correction based on seven LLMs, assuming that it is unlikely all models would be deemed reliable in the intra-rater step. This reflects the practical limit a business practitioner might explore when comparing models. This conservative design supports robust reliability assessment while accounting for computational and financial constraints.

We expect a sequential application of our framework. Practitioners would first assess the internal consistency of the models, then select consistent models for comparison against each other. Our sample size calculation accommodates this workflow with sufficient statistical power for both evaluations.

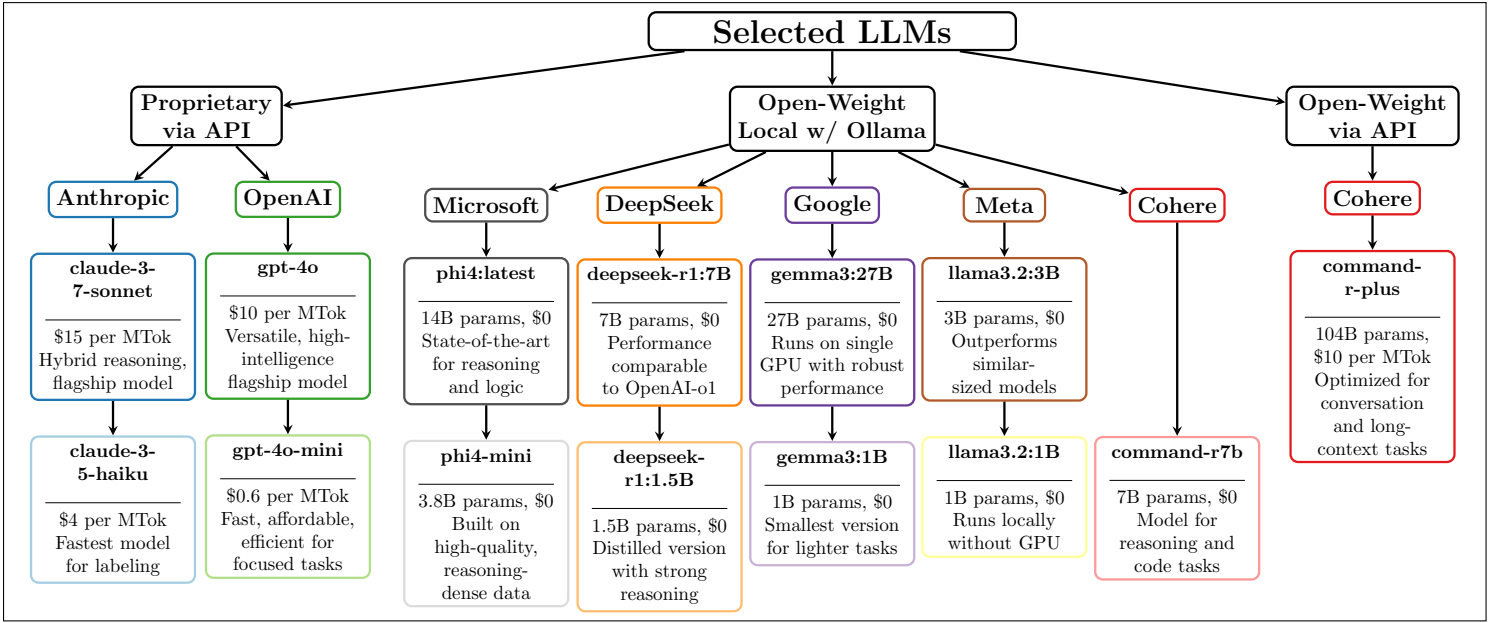
\begin{figure}[htb!]
\centering

\begin{adjustwidth}{-0.75in}{-0.75in}
\fbox{
\resizebox{1.2\textwidth}{!}{   
\begin{tikzpicture}[
    node distance=0.25in and 1in,
    every node/.style={font=\normalsize},
    level0/.style={draw, rounded corners, text width=6.5cm, align=center, font=\bfseries\Large, very thick},
    level1/.style={draw, rounded corners, text width=3.5cm, align=center, minimum height=0.6cm, very thick, font=\bfseries},
    level2/.style={draw, rounded corners, minimum width=1.5cm, align=center, minimum height=0.6cm, very thick, font=\bfseries},
    level3/.style={draw, rounded corners, align=center, text width=2.75cm, minimum height=1.2cm, very thick, font=\footnotesize},
    arrow/.style={->, >=stealth, very thick}
]

% Level 0: Main Title
\node[level0] (llms) {Selected LLMs};

% Level 1: Categories
\node[level1, below left=of llms, xshift=-1.65in, text width=2.5cm] (proprietary) {Proprietary via API};
\node[level1, below=of llms] (openollama) {Open-Weight Local w/ Ollama};
\node[level1, below right=of llms, xshift=.95in, text width=2.9cm] (openapi) {Open-Weight via API};

% Connect Level 0 to Level 1
\draw[arrow] (llms) -- (proprietary);
\draw[arrow] (llms) -- (openapi);
\draw[arrow] (llms) -- (openollama);

% Level 2: Providers under Proprietary
\node[level2, draw=cb2, below left=of proprietary, xshift=1.35in] (anthropic) {Anthropic};
\node[level2, draw=cb4, below right=of proprietary, xshift=-1.35in] (openai) {OpenAI};

% Level 2: Providers under Open API
\node[level2, draw=cb6, below=of openapi] (cohereapi) {Cohere};

% Level 2: Providers under Open Ollama
\node[level2, draw=cb10, below=of openollama] (google) {Google};
\node[level2, draw=cb8, left=of google, xshift=0.55in] (deepseek) {DeepSeek};
\node[level2, draw=cb14, left=of deepseek, xshift=0.65in] (microsoft) {Microsoft};

\node[level2, draw = cb12, right=of google, xshift=-0.4in] (meta) {Meta};
\node[level2, draw=cb6, right=of meta, xshift=-0.4in] (cohereollama) {Cohere};

% Connect Level 1 to Level 2
\draw[arrow] (proprietary) -- (anthropic);
\draw[arrow] (proprietary) -- (openai);
\draw[arrow] (openapi) -- (cohereapi);
\draw[arrow] (openollama) -- (cohereollama);
\draw[arrow] (openollama) -- (deepseek);
\draw[arrow] (openollama) -- (google);
\draw[arrow] (openollama) -- (meta);
\draw[arrow] (openollama) -- (microsoft);

% Level 3: Models with embedded rationales
\node[level3, draw=cb2, below=of anthropic] (claude37) {
    \textbf{claude-3-7-sonnet} \\
    \rule{2.5cm}{0.2pt} \\
    \$15 per MTok \\
    Hybrid reasoning, flagship model
};

\node[level3, draw=cb1, below=of claude37] (claude35) {
    \textbf{claude-3-5-haiku} \\
    \rule{2.5cm}{0.2pt} \\
    \$4 per MTok \\
    Fastest model for labeling
};

\node[level3, draw=cb4, below=of openai] (gpt4o) {
    \textbf{gpt-4o} \\
    \rule{2.5cm}{0.2pt} \\
    \$10 per MTok \\
    Versatile, high-intelligence flagship model
};

\node[level3, draw=cb3, below=of gpt4o] (gpt4omini) {
    \textbf{gpt-4o-mini} \\
    \rule{2.5cm}{0.2pt} \\
    \$0.6 per MTok \\
    Fast, affordable, efficient for focused tasks
};

\node[level3, draw=cb6, below=of cohereapi] (commandrplus) {
    \textbf{command-r-plus} \\
    \rule{2.5cm}{0.2pt} \\
    104B params, \$10 per MTok \\
    Optimized for conversation and long-context tasks
};

\node[level3, draw=cb5, below=of cohereollama, yshift=-1.3in] (commandr7b) {
    \textbf{command-r7b} \\
    \rule{2.5cm}{0.2pt} \\
    7B params, \$0 \\
    Model for reasoning and code tasks
};

\node[level3, draw=cb8, below=of deepseek] (deepseekr1) {
    \textbf{deepseek-r1:7B} \\
    \rule{2.5cm}{0.2pt} \\
    7B params, \$0 \\
    Performance comparable to OpenAI-o1
};

\node[level3, draw=cb7, below=of deepseekr1] (deepseekr1mini) {
    \textbf{deepseek-r1:1.5B} \\
    \rule{2.5cm}{0.2pt} \\
    1.5B params, \$0 \\
    Distilled version with strong reasoning
};

\node[level3, draw=cb10, below=of google] (gemma27) {
    \textbf{gemma3:27B} \\
    \rule{2.5cm}{0.2pt} \\
    27B params, \$0 \\
    Runs on single GPU with robust performance
};

\node[level3, draw=cb9, below=of gemma27] (gemma1) {
    \textbf{gemma3:1B} \\
    \rule{2.5cm}{0.2pt} \\
    1B params, \$0 \\
    Smallest version for lighter tasks
};

\node[level3, draw=cb12, below=of meta] (llama32_3) {
    \textbf{llama3.2:3B} \\
    \rule{2.5cm}{0.2pt} \\
    3B params, \$0 \\
    Outperforms similar-sized models
};

\node[level3, draw=cb11, below=of llama32_3] (llama32_1) {
    \textbf{llama3.2:1B} \\
    \rule{2.5cm}{0.2pt} \\
    1B params, \$0 \\
    Runs locally without GPU
};

\node[level3, draw=cb14, below=of microsoft] (phi4) {
    \textbf{phi4:latest} \\
    \rule{2.5cm}{0.2pt} \\
    14B params, \$0 \\
    State-of-the-art for reasoning and logic
};

\node[level3, draw=cb13, below=of phi4] (phi4mini) {
    \textbf{phi4-mini} \\
    \rule{2.5cm}{0.2pt} \\
    3.8B params, \$0 \\
    Built on high-quality, reasoning-dense data
};

% Connect Level 2 to Level 3
\draw[arrow] (anthropic) -- (claude37);
\draw[arrow] (claude37) -- (claude35);
\draw[arrow] (openai) -- (gpt4o);
\draw[arrow] (gpt4o) -- (gpt4omini);
\draw[arrow] (cohereapi) -- (commandrplus);
\draw[arrow] (cohereollama) -- (commandr7b);
\draw[arrow] (deepseek) -- (deepseekr1);
\draw[arrow] (deepseekr1) -- (deepseekr1mini);
\draw[arrow] (google) -- (gemma27);
\draw[arrow] (gemma27) -- (gemma1);
\draw[arrow] (meta) -- (llama32_3);
\draw[arrow] (llama32_3) -- (llama32_1);
\draw[arrow] (microsoft) -- (phi4);
\draw[arrow] (phi4) -- (phi4mini);

\end{tikzpicture}
}
}
\end{adjustwidth}

\caption{Selected proprietary and open-weight LLMs. Within an LLM developer, the top row indicates more performative models. The cost is per million output tokens (MTok). We provide each model's cost and features according to their provider's description.}
\label{fig:llm_selection}
\end{figure}

\paragraph*{LLM Selection} We selected 14 LLMs (depicted in Figure \ref{fig:llm_selection}) for evaluation. Our selection spans seven prominent LLM providers: \texttt{Anthropic}, \texttt{OpenAI}, \texttt{Google}, \texttt{Microsoft}, \texttt{DeepSeek}, \texttt{Meta}, and \texttt{Cohere}. We evaluated two models per provider, allowing a direct comparison of performance and cost-effectiveness within each ecosystem. This design enabled us to assess whether higher-tier models offer measurable advantages over their smaller, less expensive counterparts (i.e., is there a trade-off between compute cost and classification performance).

We varied parameter sizes among open-weight models to assess performance across scales—\\including edge models (1–3B), widely used 7B models (GPU-compatible), and mid-sized models up to 27B. We also examined the open-weights 104B parameter \texttt{command-r-plus} model via the Cohere API. Our \texttt{NVIDIA RTX 5000 Ada} with \texttt{32GB VRAM} GPU constrained us from efficiently testing models exceeding 32B parameters (locally). Furthermore, cost considerations limited our API usage to \texttt{Anthropic}, \texttt{OpenAI}, and \texttt{Cohere}.

\paragraph*{Prompt Design} Our prompt design for the case study builds upon the manual CoT and few-shot learning approaches discussed in Section \ref{sec:relwork}. Following standard practice, we developed a two-part prompt structure: (a) a system prompt with instructions and examples and (b) a user prompt that delivers the specific news article information. In the system prompt, we direct the LLM to first read the article's title and text. Then, we instruct the LLM to analyze the impact of this information on the company and identify key factors that may influence investor sentiment. The LLM is further instructed to assess the overall tone and its potential impact on the stock price. We also present it with definitions of what we mean by ``positive'' and ``negative'' sentiment and an output format that we hope it follows to facilitate our parsing of information. 
% Note that we did not require the LLM to return the results in a JSON format, as this is not supported across the 14 LLMs examined in the study. 
Our system prompt is completed with two carefully constructed examples, where we demonstrate the expected reasoning process and output format. The user prompt template passes the news article's title, full text, and stock ticker symbol to the LLM. To ensure fair comparisons, the prompt structure, including the system instructions, worked examples, and user template, was kept fixed across all 14 LLMs, and no model- or API-specific adaptations were required.

\subsection{Data Collection Phase}
 
\paragraph*{Example curation} We extracted stock news articles using the \textit{StockNewsAPI} \citep{StockNewsAPI} to investigate LLM reliability in financial sentiment classification. Data was retrieved on March 14, 2025, at approximately 12:30 EDT. Our API query used multiple parameters to obtain a pertinent dataset. We requested articles from January 05, 2025, to March 13, 2025. We selected this recent time frame to ensure these articles were not included in any LLM training data, as all evaluated models have knowledge cut-off dates before 2025. We limited the results to tickers on the NYSE and NASDAQ exchanges and US-based results. This focus on major US exchanges ensures market liquidity, standardized reporting practices, and comprehensive news coverage. We queried only positive and negative sentiment articles to maximize separation between the dichotomous sentiment classes. We collected article-type content (i.e., no videos), set 100 items per page, and iterated through multiple pages to gather our dataset.  

The initial dataset contained 10,000 articles with attributes: article URL, headline, full-text content, publishing source, publication date, related topics, API-provided sentiment classification, content type (all values = ``article''), and associated stock ticker symbols. These 10,000 articles contained 7,088 positive and 2,912 negative samples. The API labeled these sentiment tags through keyword analysis of each article's title and subtitle \citep[per their FAQ page]{StockNewsAPI}. Despite this straightforward approach, these labels are widely used as reference values in the financial sentiment analysis literature. Hence, in our case study, we consider the API-provided sentiment labels as the ``benchmark'' for our evaluation. 

We performed several preprocessing steps on the initial dataset. First, we removed articles associated with multiple tickers to ensure that each article corresponds to a single company. Next, within each sentiment category (positive and negative), we enforced ticker-level uniqueness by retaining at most one article per ticker. This prevents multiple occurrences of the same ticker within a sentiment class and avoids overweighting companies with higher news volume. Under this design, a ticker may appear once in the positive set and once in the negative set, yielding a maximum of two occurrences across the preprocessed dataset.
% retain only news focused on a single stock. This ensures sentiment classification applies to specific companies rather than mixing signals across multiple entities. We implemented a sampling strategy that prevents the same ticker from appearing multiple times within the same sentiment category. Thus, a ticker can appear at most twice in our preprocessed dataset. 
This was followed by random downsampling of the dataset to 1,350 articles, equally divided between positive and negative. Our final dataset for LLM-based classification contained 1,183 unique stock tickers. 
% Example tickers include American Airlines (AAL), Apple (AAPL), Goldman Sachs (GS), and Johnson \& Johnson (JNJ). This diversity reflects the composition of stocks within the NYSE and NASDAQ. 

\paragraph*{Multi-LLM-based sentiment classification with repeats}

Our case study examined sentiment classification across $n=1,350$ financial news articles using a diverse set of 14 LLMs (from seven providers). Each model processed each article five times, yielding $6,750 \ (1,350 \times 5)$ prompt completions per LLM. To ensure consistency, we constructed prompts by combining the article title, full text, and ticker symbol with a standardized system message. The model suite encompassed commercial API-based models and locally hosted open-weight alternatives deployed via \texttt{Ollama}. We configured all runs with a set temperature of 0 to minimize the models' creativity. However, our proposed framework is flexible and allows users to adjust the temperature according to their application needs, including experiments that assess how model consistency changes under different temperature settings. Furthermore, we set a 3000-token limit to ensure all models have the same output configurations. We implemented a retry mechanism for up to three failed attempts per request in case of any API connection and rate-limiting issues. For sentiment extraction, we used regular-expression-based pattern-matching techniques to extract the binary classification labels (``positive'' or ``negative'') and developed specialized logic to resolve inconsistent response formatting. 
% We maintained comprehensive logging of all outputs and associated metadata throughout the process for traceability and experimental verification. 

\begin{figure}[htb!]
\centering
\begin{adjustwidth}{-0.65in}{-0.65in}
\begin{subfigure}[b]{0.49\linewidth}
\centering
\includegraphics[width=0.99\textwidth, frame]{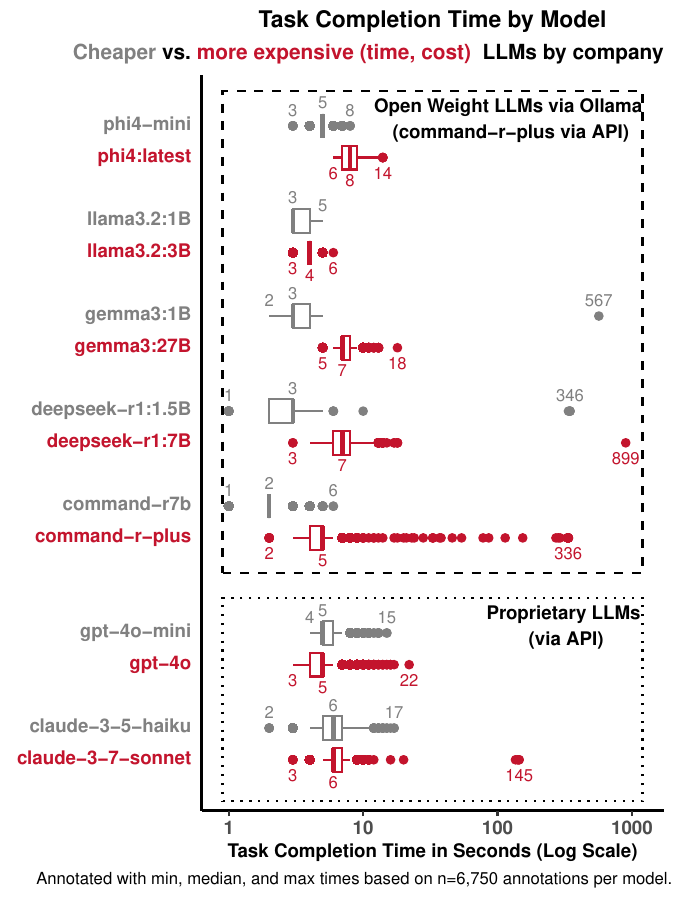}
\caption{LLM inference time distribution.}
\label{tab:inference-times}
\end{subfigure}
\begin{subfigure}[b]{0.49\linewidth}
\centering
\includegraphics[width=0.99\textwidth, frame]{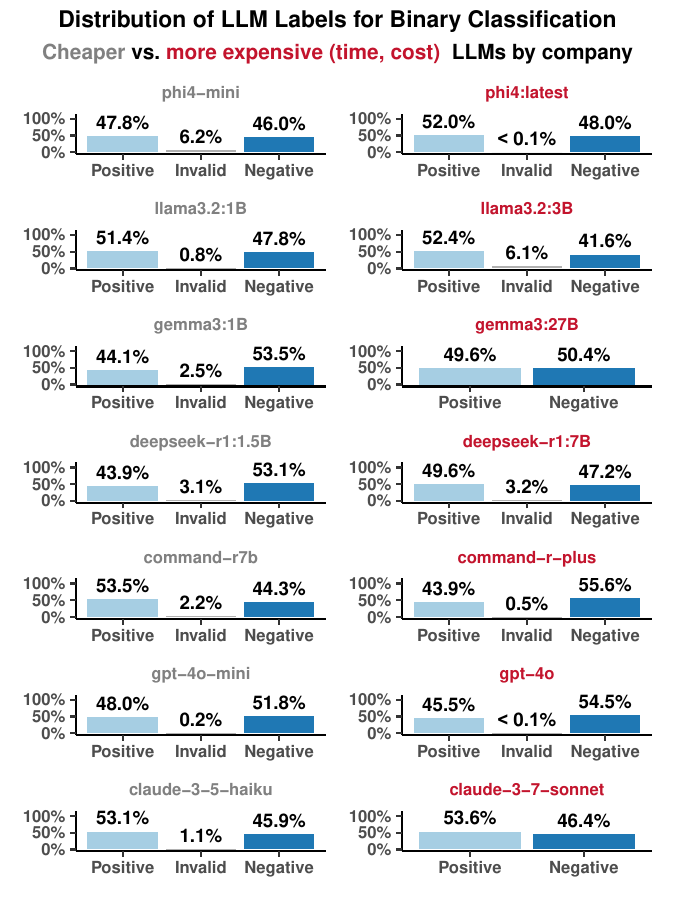}
\caption{Distribution of sentiment labels per LLM.}
\label{fig:label-dist}
\end{subfigure}
\end{adjustwidth}

\vspace{-0.64\baselineskip}

\caption{Comparison of LLM inference times and generated sentiment distributions. An ``invalid'' label is assigned when the model output cannot be mapped to the required ``positive'' or ``negative'' categories. For example, the model responds with terms like ``neutral'' or ``unsure,'' or when the output cannot be parsed into either label.} \label{fig:llm_comparison}
\end{figure}

Our analysis of inference latency and label distribution reveals significant performance variations across model types. Figure~\ref{tab:inference-times} presents a detailed summary of inference latency performance across models. API-based models demonstrate relatively consistent performance with median inference times between 5-6 seconds, while locally hosted models show parameter-dependent variability. Smaller models such as \texttt{command-r7b} (7B parameters) and \texttt{gemma3:1B} execute faster with 2-4 second median times. Conversely, larger models such as \texttt{gemma3:27B} require approximately 7 seconds per inference. Reasoning-\texttt{deepseek-r1} models exhibit significantly higher maximum latencies (346-899 seconds) than their counterparts due to occasional extended reasoning loops that substantially increase processing time. We also saw relatively large outliers in some API-based models, with \texttt{claude-3-7-sonnet} and \texttt{command-r-plus} showing elevated maximum times of 145s and 336s, respectively. These outliers stem from a combination of our retry mechanism (which incorporated a 30-second delay between attempts), network variability, server-side reasoning loops, and load balancing in API services during peak usage periods. Figure~\ref{fig:label-dist} shows the distribution of labels per model. Since our original dataset was balanced according to the \textit{StockNewsAPI} sentiment labels, we expected an even distribution between positive and negative classifications. In Figure~\ref{fig:label-dist}, the ``invalid'' label represents the inconsistency between our prompt (which explicitly instructed the LLM to restrict its classification to either ``positive'' or ``negative'') and the generated classification, which sometimes included terms such as ``neutral'' or ``unsure.''

Across all models, we observed 39,431 (48.68\%) positive labels, 39,648 (48.95\%) negative labels, and 1,921 (2.37\%) invalid labels. The approximately even distribution of ``positive'' and ``negative'' labels is consistent with the \textit{StockNewsAPI} sentiment labels. At the LLM level, Figure~\ref{fig:label-dist} shows notable deviations in the label distribution. \texttt{claude-3-7-sonnet} (53.6\% positive, 46.4\% negative) and \texttt{command-r7b} (53.5\% positive, 44.3\% negative) have a modest positive bias. Conversely, \texttt{gpt-4o} (45.5\% positive, 54.5\% negative) and \texttt{command-r-plus} (43.9\% positive, 55.6\% negative) exhibit a negative sentiment tendency. The inconsistency rates vary, with generally more ``invalid'' labels associated with the \textit{cheaper} models. For example, only \texttt{claude-3-7-sonnet} and \texttt{gemma3:27B} had no ``invalid'' labels. On the other hand, their \textit{cheaper} counterparts had 1.1\% and 2.5\% of their respective classifications labeled as ``invalid''.

\subsection{Reliability Assessment}

\begin{figure}[htb!]
\centering
\begin{adjustwidth}{-0.65in}{-0.65in}
\begin{subfigure}[b]{0.49\linewidth}
\centering
\includegraphics[width=0.99\textwidth, frame]{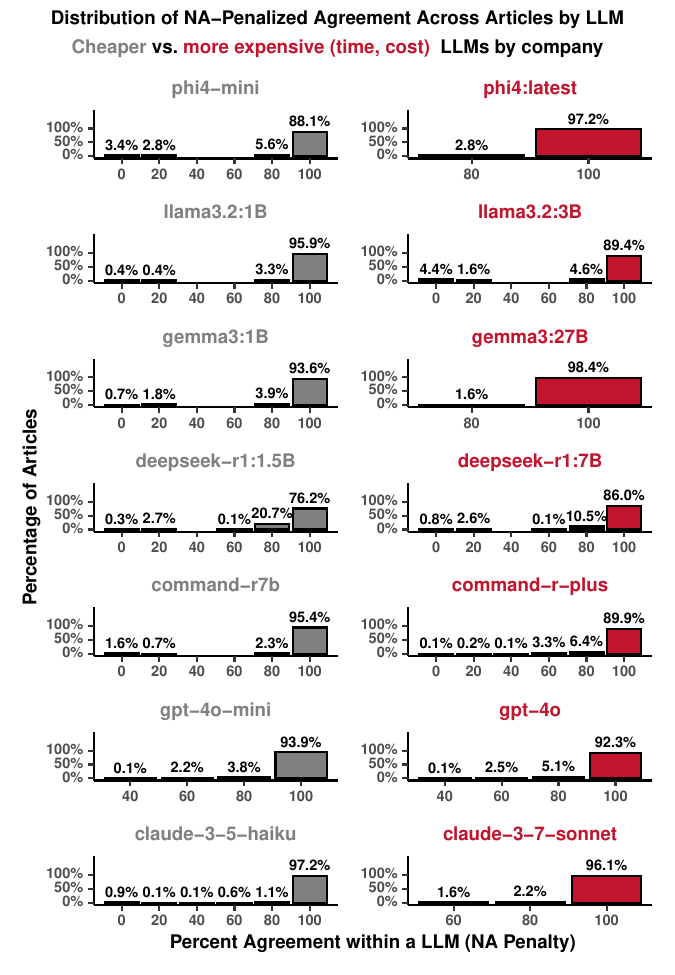}
\caption{NA-penalized agreement by model}
\label{fig:na_pen_agreement}
\end{subfigure}
\begin{subfigure}[b]{0.49\linewidth}
\centering
\includegraphics[width=0.99\textwidth, frame]{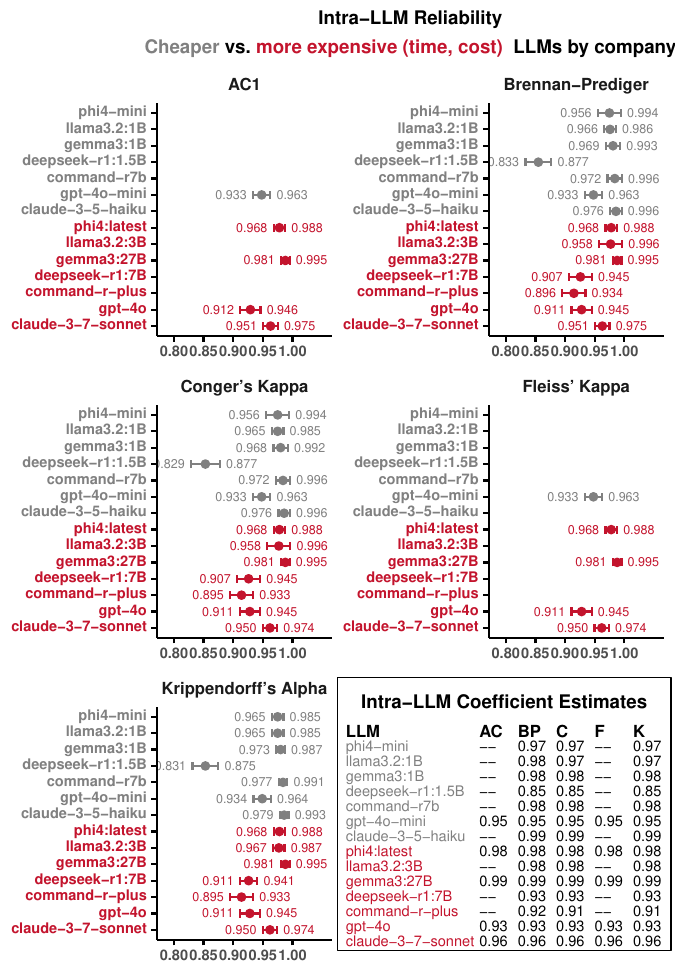}
\caption{Intra-LLM reliability coefficient estimates}
\label{fig:intra_stats}
\end{subfigure}
\end{adjustwidth}

\vspace{-0.64\baselineskip}

\caption{The distribution of the intra-LLM reliability coefficient estimates. \texttt{AC1}, and \texttt{Fleiss' Kappa} cannot be computed when all five raters (replicates) provide \texttt{NA} (i.e., ``invalid'') labels for a given news article. Hence, models with only invalid labels for certain items have no values for these metrics in the dot-plot with error bars and the coefficient estimate tables (coefficient names were abbreviated for space considerations).} \label{fig:intra}
\end{figure}

\paragraph*{Intra-rater reliability} Figure~\ref{fig:intra} presents our intra-LLM reliability results. Figure~\ref{fig:na_pen_agreement} shows the \texttt{NA}-penalized agreement distributions for each model, highlighting the percentage of articles where all five replicates generated identical classifications. The \texttt{NA}-penalty reflects how we count agreement in the presence of invalid responses, where \texttt{NA}s are ignored in the numerator but are kept in the denominator. For example, if an LLM returns three invalid responses and two ``positive'' labels for a given article, the percent agreement is calculated as $\frac{2 \text{ positive}}{5 \text{ replicates}} = 40\%$. Figure~\ref{fig:intra_stats} shows the chance-adjusted reliability coefficient estimates (dots) and their \v{S}id\'{a}k-adjusted confidence intervals (whiskers), constructed to maintain 90\% family-wise confidence within each cost group. Specifically, seven intervals are shown for the cheaper, less resource-intensive models and seven for the expensive alternatives.

% Together, these visualizations provide a comprehensive assessment of model-level internal consistency. Note that the \texttt{AC1} and \texttt{Fleiss' Kappa} cannot be computed when a model returns invalid responses, explaining the missing (or ``--'') values for several models.

Most models demonstrated remarkably high intra-rater consistency. The \texttt{NA}-penalized agreement distributions showed that most models achieve 100\% agreement on 88--98\% of articles, with mean agreement rates exceeding 0.93 for all models. The other reliability metrics reinforce these trends. Models such as \texttt{gemma3:27B} and \texttt{claude-3-5-haiku} achieved values about 0.99 across \texttt{Brennan-Prediger}, \texttt{Conger's Kappa}, and \texttt{Krippendorff's Alpha}. Interestingly, for our case study, the intra-LLM consistency was similar across smaller and larger models within the same provider. This suggests diminishing returns from increased model size, at least with respect to self-consistency, except in the case of the \texttt{deepseek} models. 

The \texttt{deepseek} models stood out as clear exceptions to this trend. \texttt{deepseek-r1:1.5B} achieved perfect agreement on ``only'' 76.2\% of articles (the lowest among all models) and showed a wider spread across agreement levels, including 20.7\% of stock news articles with 80\% agreement. Correspondingly, its reliability coefficients were more moderate, all at 0.85 across \texttt{Brennan-Prediger}, \texttt{Conger's Kappa}, and \texttt{Krippendorff's Alpha}, suggesting that its disagreements were more dispersed and less systematic. In contrast, the larger \texttt{deepseek-r1:7B} model achieved both a higher perfect agreement rate (86\%) and stronger reliability coefficients (0.93 for all three metrics), indicating more consistent internal reasoning and a tighter alignment between raw agreement rates and reliability-adjusted measures.

\begin{figure}[htb!]
    \centering
    \includegraphics[width=0.68\textwidth, frame]{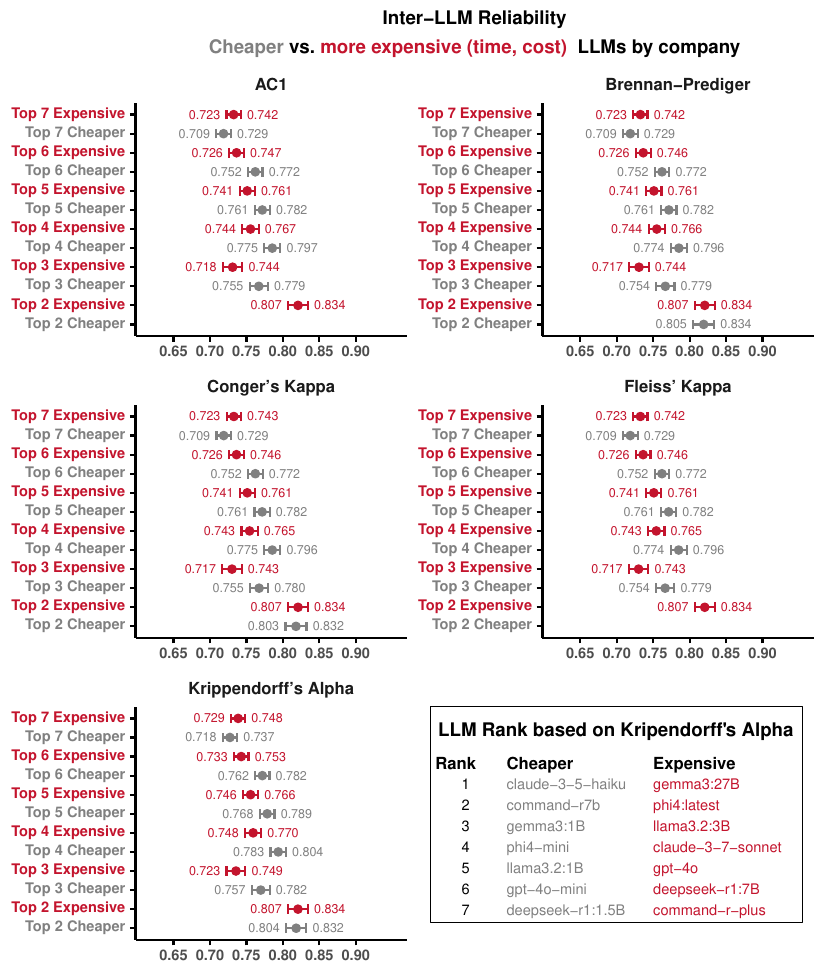}
    \caption{A dot plot with whiskers representing the 95\% confidence intervals for the inter-rater reliability metrics for different combinations of LLMs.}
    \label{fig:inter_dist}
\end{figure}

\paragraph*{Inter-rater reliability} After confirming strong internal consistency within individual models, we examine agreement between different LLMs. While all models exhibit mean reliability coefficients above 0.9 across metrics, using all 14 LLMs is unlikely in practice, so we adopt a more targeted inter-rater analysis. Models are grouped into two categories: (a) larger, more expensive models, and (b) cheaper, less resource-intensive alternatives. Instead of analyzing all possible combinations within each group, we focus on top-$N$ combinations ranked by \texttt{Krippendorff's Alpha}, with $N$ ranging from 2 to 7. For example, when $N=2$, we compare \texttt{gemma3:27B} and \texttt{phi4:latest} (larger models with $\alpha$ values of 0.988 and 0.978) to \texttt{claude-3-5-haiku} and \texttt{cammand-r7b} (cheaper models, scoring 0.986 and 0.984). This approach supports meaningful comparisons of practical model subsets without requiring exhaustive evaluation. Figure~\ref{fig:inter_dist} presents the results.   

% \begin{table}[ht]
% \setlength{\tabcolsep}{0.5pt}
% \begin{adjustwidth}{-0.5in}{-0.5in}
%     \centering
%     \caption{Inter-Rater Agreement metrics.}
%     \label{tab:inter_coeffs}
%     \footnotesize{
%     \begin{tabular}{C{2.4cm}C{2.4cm}C{2.4cm}C{2.4cm}C{2.4cm}C{2.4cm}C{2.4cm}} \toprule
%   \textbf{\% Agree NA Dropped} & \textbf{\% Agree NA Penalized} & \textbf{Conger's Kappa} & \textbf{Fleiss' Kappa}  & \textbf{AC1} & \textbf{B-P Kappa} & \textbf{Kripp Alpha}\\ \hline
%   0.9790 (0.1070) & 0.9700 (0.1320) & 0.7159 (0.0046) & 0.7158 (0.0046) & 0.7158 (0.0046) & 0.7158 (0.0046) & 0.7242 (0.0045)\\ \hline \\[-1em]
%     \end{tabular}
%     }\\
%     \footnotesize{*Standard Error is included in parentheses.}
% \end{adjustwidth}
% \end{table}

Figure \ref{fig:inter_dist} reveals two patterns. First, the addition of more LLMs generally reduces the inter-rater reliability metrics, which is an expected result. Second, reliability remains high across most combinations$\times$metrics, typically exceeding 0.70.

\subsection{Validity Analysis}

We assessed the predictive (criterion-related) validity of LLM-generated sentiment annotations against two reference standards: (1) benchmark model labels derived from the \textit{StockNewsAPI} and (2) an external market-based criterion reflecting actual stock price movements. To ensure a fair evaluation, we treated ``invalid'' labels as incorrect, since failing to classify an article violates our instruction that the LLM must provide either a ``positive'' or ``negative'' label. 

For the benchmark comparison, LLM classifications were evaluated against the API-provided sentiment tags. For the external criterion comparison, we defined success using excess return: a stock’s performance relative to the \texttt{S\&P500} index (\texttt{\^{}GSPC}) one trading day after the article's publication. A ``positive'' label was considered correct if the stock outperformed the index, while a ``negative'' label was correct if the stock underperformed. This dual approach allows us to assess whether LLM sentiment predictions align both with existing labeling heuristics and with real-world market outcomes. 

\begin{figure}[htb!]
\centering
\begin{adjustwidth}{-0.65in}{-0.65in}
\begin{subfigure}[b]{0.49\linewidth}
\centering
\includegraphics[width=0.99\textwidth, frame]{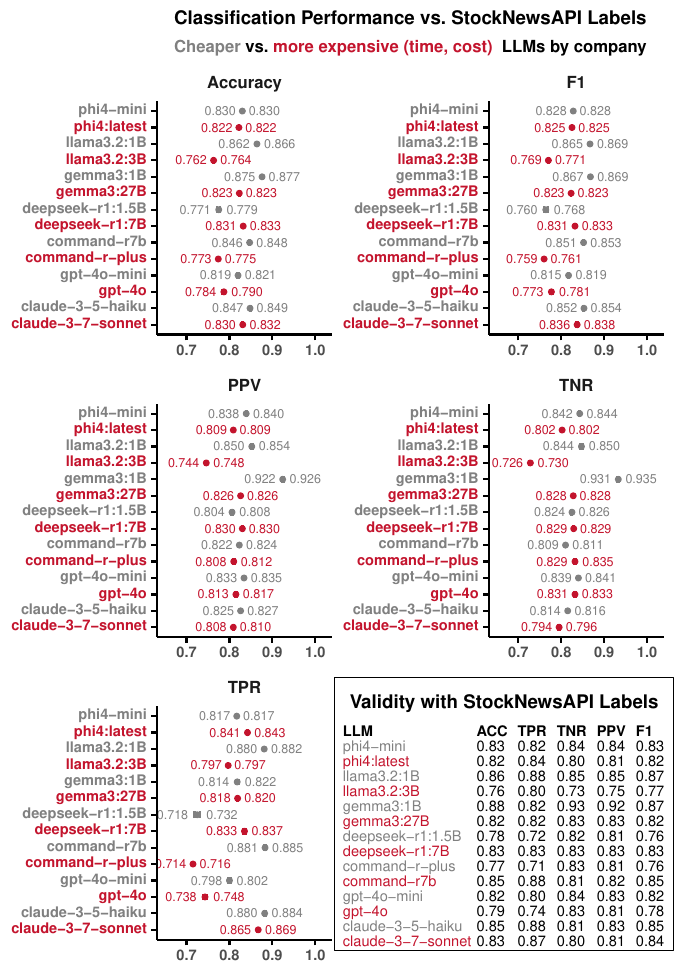}
\caption{LLM classification performance vs. benchmark model}
\label{fig:gold_standard}
\end{subfigure}
\begin{subfigure}[b]{0.49\linewidth}
\centering
\includegraphics[width=0.99\textwidth, frame]{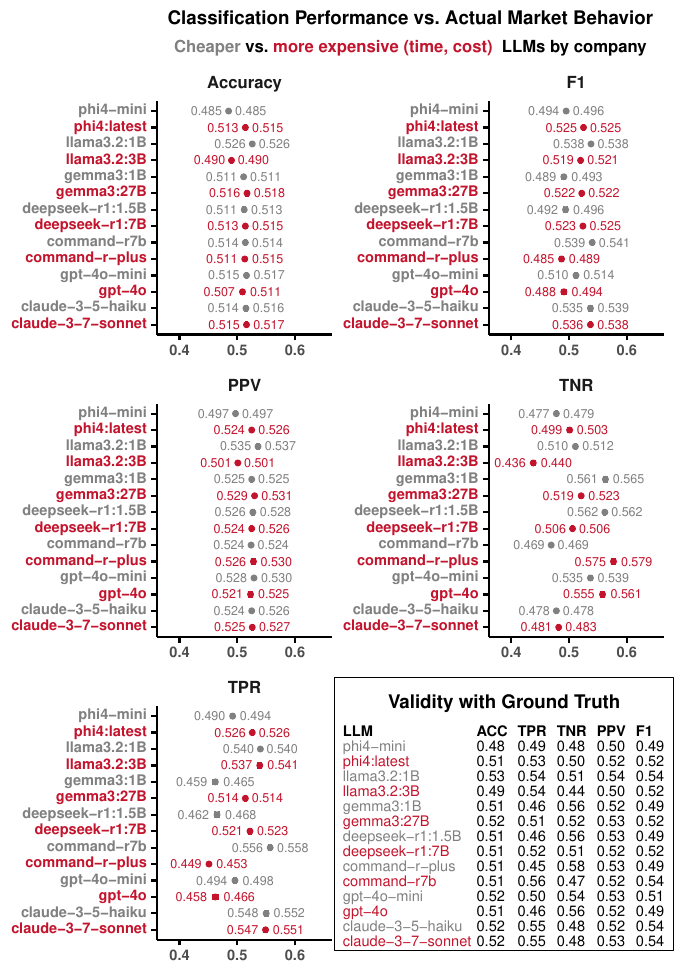}
\caption{LLM classification performance vs. external criterion}
\label{fig:ground_truth}
\end{subfigure}
\end{adjustwidth}

\vspace{-0.64\baselineskip}

\caption{Validating the LLM labels, with those obtained from the \textit{StockNewsAPI}, and the actual stock excess return compared to the \texttt{\^{}GSPC}. The dot represents each metric's mean value (from its intra-rater five replicates), and the whiskers' length reflects the standard error (as opposed to the confidence intervals in the previous charts).} \label{fig:validity}
\end{figure}

Figure~\ref{fig:validity} summarizes our assessment of predictive validity against the benchmark model and the external market-based criterion. We report accuracy (proportion of correct predictions), true positive rate (TPR, sensitivity), true negative rate (TNR, specificity), positive predictive value (PPV, precision), and F1 score (harmonic mean of precision and recall) to provide a comprehensive assessment of classification performance.

%LLMs were evaluated not only on classification accuracy but also on their ability to consistently produce sentiment predictions. We assessed whether LLM annotations align with ``gold-standard'' sentiment labels from the \textit{StockNewsAPI} and whether their predictions captured the real market behavior (``ground truth''). The ``ground truth'' has been conservatively defined using the notion of ``excess return'', which measures a stock's performance relative to a benchmark index. Specifically, we compared the stock's return to the daily percentage change of the \texttt{S\&P500} (\texttt{\^{}GSPC}) one trading day after the article's release. A ``positive'' label was considered correct if the stock delivered a positive excess return, i.e., it outperformed the index. In contrast, a ``negative'' label was correct if the stock underperformed the index. In this setup, the \texttt{\^{}GSPC} serves as a proxy for overall market performance, allowing us to evaluate if the LLMs can effectively ``beat the market.'' Figure~\ref{fig:validity} summarizes our validations with the ``gold-standard'' and ``ground truth'' labels, using standard classification accuracy metrics. We report accuracy (proportion of correct predictions), true positive rate (TPR, sensitivity), true negative rate (TNR, specificity), positive predictive value (PPV, precision), and F1 score (harmonic mean of precision and recall) to provide a comprehensive assessment of classification performance.

\paragraph*{Performance against benchmark model} When compared against the \textit{StockNewsAPI} sentiment labels, most LLMs demonstrated strong performance (see Figure~\ref{fig:gold_standard}) with mean accuracy values ranging from 0.76 to 0.88. The highest accuracy was achieved by \texttt{gemma3:1B} (0.88), followed by \texttt{llama3.2:1B} (0.86) and \texttt{claude-3-5-haiku} (0.85). Interestingly, these three smaller models had improved accuracy rates compared to their larger counterparts. For example, \texttt{llama3.2:1B} surpassed \texttt{llama3.2:3B} by 10 percentage points (0.86 vs. 0.76), and \texttt{command-r7b} outperformed \texttt{command-r-plus} (0.85 vs. 0.77).  

For most models, true positive rates (sensitivity) were generally higher than true negative rates (specificity). This suggests that LLMs were more effective at identifying positive than negative sentiment. The highest F1 score was achieved by \texttt{gemma3:1B} (0.868), followed closely by \texttt{llama3.2:1B} (0.867) and \texttt{claude-3-5-haiku} (0.85), indicating strong overall performance balancing precision and recall.

\paragraph*{Performance against the external market-based criterion} Figure~\ref{fig:ground_truth} presents model performance against the external market-based criterion. Examining the precision (PPV) results, all models performed near chance level with values ranging from 0.5 to 0.54, indicating that positive predictions were correct about half the time. Similarly, the TPR and TNR values hover around 0.50, slightly varying across models. For example, \texttt{claude-3-7-sonnet} shows the highest TPR (0.55) but a below-average TNR (0.48), while \texttt{command-r-plus} demonstrates the opposite pattern with the highest TNR (0.58) but one of the lowest TPR values (0.45). The F1 scores, which balance precision and recall, range from 0.49 (e.g., \texttt{gemma3:1B}) to 0.54 (e.g., \texttt{claude-3-7-sonnet}), but are close to 0.50 for all models. 

These results suggest that while LLMs can effectively replicate \textit{StockNewsAPI's} sentiment classification methodology, neither approach validly predicts actual market movements. This finding aligns with the \textit{efficient market hypothesis}; publicly available information is rapidly incorporated into stock prices, making predictions based solely on news sentiment challenging. We found that ensembling the LLMs did not materially improve predictive validity (see supplementary materials).

\section{Discussion}
\label{sec:disc}
Our study introduces a structured scientific framework for evaluating the reliability and validity of LLM-based binary text annotation and classification. The framework systematically assesses both the internal consistency of individual LLMs and the agreement between different models. In addition, we evaluate the predictive validity of LLM-generated annotations relative to a benchmark model and an external criterion. We demonstrate the utility and flexibility of this approach through an extensive financial sentiment classification case study.

Our evaluations revealed that LLMs generally demonstrate remarkably high intra-rater consistency, with most achieving perfect agreement on 88-98\% of examples across five replicates, despite minimal prompt engineering.  In addition, the inter-rater reliability analysis across model groups demonstrated strong reliability. Although adding models slightly decreased agreement levels, overall reliability remained high. Reliability of annotations is critical in decision support contexts such as finance or healthcare, where model outputs need to be predictable, repeatable, and trustworthy. 

A notable result of this study is that smaller models, such as \texttt{GPT-4o-mini} and \\\texttt{Claude-3-5-haiku}, demonstrated performance metrics comparable to larger, more expensive models such as \texttt{GPT-4o} and \texttt{Claude-3-7-sonnet} for our financial news sentiment annotation task. This challenges the “fear of missing out” (FOMO) mentality that drives many organizations to continually upgrade to newer, larger models without clear evidence of improved task performance. This finding is particularly relevant for firms seeking to scale LLM solutions while managing computational and financial costs. The prevailing industry trend towards ever larger models has often overshadowed the potential of smaller, faster models to deliver high-quality results, especially for narrowly defined tasks like binary classification \citep{chiang2023can}. Our observations align with prior studies questioning the cost-effectiveness of larger models for routine classification tasks \citep{reiss2023testing}. These results reinforce the importance of selecting models based on task-specific performance evaluations rather than assumptions based on model size.

Meanwhile, although many tested LLMs—including \texttt{gemma3:1B}, \texttt{llama3.2:1B}, and \newline\texttt{Claude-3-5-haiku}—achieved high levels of accuracy against the benchmark model, their annotations did not align as strongly with real-world financial market movements (i.e., the external criterion). This discrepancy underscores that while LLMs can produce semantically consistent labels, these labels may not always reflect actual behavioral outcomes such as stock price changes. This finding echoes prior research (e.g., Aguda et al. \cite{aguda2024large}) suggesting that LLMs may excel at tasks requiring semantic alignment but struggle with tasks requiring complex causal inference or behavioral prediction. As such, LLM-generated annotations may be more appropriate for descriptive or classification-oriented applications than for predictive modeling, unless additional task-specific tuning is applied.

It is important to note that both validity references used in our case study are imperfect ground truths. The API labels, while commonly used in financial sentiment research, are generated using rule-based keyword heuristics and may not fully capture nuanced investor sentiment. Likewise, short-horizon market returns reflect a complex mixture of information, noise, and macroeconomic dynamics, which limits their ability to serve as a definitive behavioural criterion. These limitations do not undermine the proposed framework; rather, they illustrate a central challenge in LLM validation studies, where reference labels and external criteria are themselves subject to measurement error. Our framework is designed to accommodate alternative or high-quality validity references in future applications, such as expert-annotated datasets or multi-day event study outcomes. Our case study demonstrates how validity assessment can be carried out even when ground truths are imperfect.

In sum, while careful experimental design, appropriate sample size planning, and nuanced interpretation of performance metrics remain critical, as emphasized by \cite{gilardi2023chatgpt}, researchers and practitioners must also carefully distinguish between tasks involving semantic judgment and those requiring true predictive inference.

\section{Theoretical and Practical Implications}
\label{sec:implications}
\subsection{Theoretical Implications}

This study advances the theoretical foundations of LLM reliability evaluation by introducing a scientific framework for using LLMs as structured raters in classification tasks. First, it contributes to growing work on replicable LLM systems by proposing a statistical method for determining minimum sample sizes for intra-rater reliability, addressing a gap where prior studies relied on arbitrary choices (e.g., \cite{zhou2021breaking}). This psychometrically informed approach bridges empirical LLM evaluation with reproducible social science methods. It shifts the focus from the accuracy of the model \citep{kazari2025scaling} to the stability of the measurement, viewing LLMs as bounded probabilistic annotators. This perspective aligns the evaluation of LLMs with long-standing measurement theory practices that emphasize consistency, validity, and reproducibility. Second, the empirical findings challenge the assumption that larger models offer greater reliability. Smaller models (e.g., \texttt{llama3.2:1B}, \texttt{claude-3-5-haiku}) matched or exceeded larger ones in consistency and accuracy, suggesting model architecture, prompt adherence, and task specificity may matter more than scale. This invites refinement of LLM scaling theories, proposing that emergent abilities may not extend uniformly to structured classification tasks. Third, the disconnect between high consistency and low predictive validity highlights a theoretical boundary. While consistency is necessary for reliability, it is not sufficient. Highly consistent models like \texttt{Gemma 3} and \texttt{Claude 3} failed to predict stock movements, supporting prior claims about LLMs’ domain-specific limitations \citep{tornberg2023chatgpt, stureborg2024large}. This underscores the need to distinguish tasks requiring semantic consistency from those requiring real-world prediction, avoiding overgeneralization across fundamentally different task types. Finally, the framework’s modular design supports its application across domains such as legal text labeling, medical triage, and educational content tagging. As LLMs gain traction in industry, structured evaluations are critical for user trust, consistent with findings from human-AI collaboration research \citep{liu2024poliprompt}.

\subsection{Practical Implications}

This study offers actionable guidance for practitioners applying LLMs to classification and annotation tasks. First, the proposed framework offers a replicable, resource-efficient method for validating LLMs before full deployment. By adapting psychometric reliability analyses, practitioners can establish empirically justified consistency thresholds and benchmark models without relying on subjective evaluations. Second, smaller, less expensive LLMs proved highly viable, achieving reliability and accuracy comparable to larger models. With small models offering 5–29 times cost reductions \citep{irugalbandara2024scaling}, organizations in resource-constrained environments can prioritize task-specific suitability over defaulting to flagship or high-parameter models. Third, the framework emphasizes the need to handle invalid responses explicitly—using \texttt{NA}-penalized and \texttt{NA}-dropped strategies—to ensure reliability assessments reflect real-world conditions. Practitioners can apply these approaches to build more robust, production-ready annotation pipelines. 
Fourth, while LLM annotations reliably replicate benchmark labels, their failure to predict future outcomes highlights a key limitation. LLMs are well-suited for structured classification tasks, but should not be assumed to provide predictive insights into stochastic real-world phenomena without additional modeling layers or complementary methods. Finally, the open-source code and replicable experimental setup enable rapid adoption and adaptation by research groups and industry teams, promoting best practices in evaluating LLM-based systems for annotation and decision support. Together, these insights offer a practical roadmap for organizations seeking to deploy LLMs responsibly, efficiently, and with measurable reliability.

\section{Limitations and Future Research}
\label{sec:limitations}
While this study presents a rigorous framework for evaluating LLM consistency, several limitations suggest directions for future research. First, our analysis focused on binary classification, whereas many real-world tasks involve multiclass settings. Extending the framework to multiclass contexts introduces challenges, such as adapting reliability metrics and recalibrating sample size estimates for increased label complexity. An additional direction for future research involves extending the framework to fuzzy or graded annotation settings, where LLMs provide confidence scores or degrees of membership rather than discrete class labels. Generalizability is limited by the case study’s focus on financial sentiment. Applying the framework across domains will clarify whether observed reliability patterns hold more broadly and how model size influences performance in varied tasks. Third, handling non-compliant LLM responses remains a challenge. We used a lenient regular-expression approach to extract misformatted labels; alternatives include penalizing non-conformity or constraining outputs to formats like JSON \citep{guzman2024introduction}. Finally, the framework currently applies only to text-based tasks. Future work should explore its use with non-text modalities, such as image classification via vision transformers, to assess whether reliability patterns extend to multimodal systems—particularly relevant for Industry 4.0 and 5.0 applications \citep{megahed2025adapting}.

\section{Conclusion}
\label{sec:conclusion}
LLMs are increasingly used for text annotation and classification in decision support systems, where rigorous evaluation is critical. This study presents a four-phase framework for assessing LLM consistency, reliability, and validity in binary classification, grounded in psychometric principles for model selection, sample sizing, and reliability analysis. In a financial sentiment case study involving 14 LLMs and 1,350 articles, both small and large models achieved high intra- and inter-rater reliability, with smaller models often matching or exceeding larger ones in consistency and cost-efficiency. This finding can help organizations optimize resources. However, while LLMs replicated benchmark sentiment labels, they failed to predict stock movements, underscoring the gap between annotation reliability and real-world predictive validity. LLMs are well-suited for structured annotation but require careful evaluation in predictive contexts. The proposed framework offers replicable tools for deploying LLMs in scalable, trustworthy annotation workflows.

\section*{Online Supplementary Materials}
Our knitted \faRProject \ Markdown, available at \url{https://fmegahed.github.io/research/llm_consistency/llm_consistency.html}, integrates the \faRProject \ and Python code to evaluate LLM consistency in binary text annotation tasks. It includes the code and the corresponding results, such as figures and tables, providing a self-contained resource for implementing our framework.

To facilitate reproducibility and encourage further research in this area, we provide comprehensive materials in our GitHub repository \url{https://github.com/fmegahed/llm_consistency}. The repository contains four main components:
\begin{enumerate}[label=(\arabic*), nosep]
    \item a \emph{data} folder with the stock new articles and their labels.
    \item a \emph{rmarkdown} folder containing our \faRProject \, Markdown.
    \item a \emph{figure} folder storing code to compile the figures and results in this paper; and
    \item a \emph{results} folder storing all experimental outputs.
\end{enumerate}

\section*{Acknowledgments}

This research was supported by US Bank through an unrestricted technology fund, and the \textit{Farmer School of Business}, through a \textit{Summer Research Grant} and the \textit{Van Andel Endowed Professorship} in Business Analytics. 
% These contributions provided the computational resources and monetary support necessary to conduct the experiment. 
Note that US Bank did not participate in the study design and the performed research.

% \linespread{1} \selectfont
\label{sec:ref}
\bibliographystyle{abbrvnat}
\bibliography{refs}

\end{document}